\documentclass{article}
     \PassOptionsToPackage{numbers, compress}{natbib}

     \usepackage[preprint]{neurips_2019}

\usepackage[utf8]{inputenc} 
\usepackage[T1]{fontenc}    
\usepackage{hyperref}       
\usepackage{url}            
\usepackage{booktabs}       
\usepackage{amsfonts}       
\usepackage{nicefrac}       
\usepackage{microtype}      
\usepackage{graphicx}
\usepackage{amsmath}
\usepackage{amssymb}
\usepackage{mathtools}
\usepackage{algorithm,algorithmic}
\usepackage{subfig}

\title{Interpretable Few-Shot Learning \\ via Linear Distillation}

\author{%
  Arip Asadulaev\\
  ITMO University\\
  Saint-Petersburg, Russia \\
  \texttt{aripdotcom@mail.ru} \\
  \And
  Igor Kuznetsov\\
  ITMO University\\
  Saint-Petersburg, Russia\\
  \texttt{igorkuznetsov14@gmail.com} \\
  \And 
  Andrey Filchenkov\\
  ITMO University\\
  Saint-Petersburg, Russia\\
  \texttt{afilchenkov@itmo.ru} \\
}

\begin{document}

\maketitle

\begin{abstract}
It is important to develop mathematically tractable models than can interpret knowledge extracted from the data and provide reasonable predictions. In this paper, we present a Linear Distillation Learning, a simple remedy to improve the performance of linear neural networks. Our approach is based on using a linear function for each class in a dataset, which is trained to simulate the output of a \textit{teacher} linear network for each class separately. We tested our model on MNIST and Omniglot datasets in the Few-Shot learning manner. It showed better results than other interpretable models such as classical Logistic Regression.

\end{abstract}

\section{Introduction}

One of the topical issues in the machine learning domain is the interpretability of models. For sophisticated models such as ensemble methods or deep neural networks (DNN), it is not easy to understand the rationale behind their predictions. The increasing adoption of machine learning in a variety of real-world systems contributes to the heightened into models understanding. Many papers proposing different techniques intended to improve model interpretability~\cite{Visualint,BBM} and measured the effect of different interpretability methods on user trust, ability to simulate models, and ability to detect mistakes~\cite{MMMI}. A taxonomy for categorizing interpretability methods with different properties was presented in Lipton notes~\cite{Lipton}. 

Some papers seek to discuss how machine learning researchers interpret models by measuring the interpretability of the model on a specific tasks~\cite{Miller}.  

In this paper, we approach the interpretability problem by creating a simple few-shot learning-based method using linear neural networks (LNNs) that are feedforward neural networks with no nonlinearities. Output of each layer is multiplication of its weights to its input and the output of such network can be computed by matrix multiplication of all weight matrices.


The main advantage of LNNs is that they are tractable and interpretable. Linear models learn monotonic relationships between the features and the target. They have been used for a long time by statisticians, computer scientists and widespread in academic fields such as medicine, sociology, psychology, and many more quantitative research fields. In these areas, it is important to not only predict, e.g., the clinical outcome of a patient but also to explain the model predictions. The linearity of the learned relationship makes the interpretation easy. 
The estimation procedure of linear model performance is straightforward and weights in the linear equations have an easy to understand interpretation. 
%
Despite LNNs are trivial from a representational perspective, the process of training such networks is in research focus. The study of linear networks loss dynamics is a source of novel research questions, helpful insights and brand new ways of looking at certain aspects of deep learning~\cite{DBLP:conf/nips/Kawaguchi16,DBLP:journals/corr/LuK17,DBLP:journals/corr/SaxeMG13,DBLP:journals/jmlr/SrivastavaHKSS14,DBLP:conf/icml/SutskeverMDH13}.

Deep neural networks tend to operate on raw features and learn rich representations that can be visualized, verbalized, or used for further processing. Increasing depth of LNN does not increase its expressive power, and in order to get comparable performance, linear models often must operate on heavily hand-engineered features. 

In this paper, we were guided by a research question \textbf{If one can improve linear network performance without harming its interpretability property?} We claim that we found the positive answer to this question.

We present a Linear Distillation Learning (LDL), a method that uses a linear function for each class in dataset, which is trained to simulate output of some \textit{teacher} linear function for each class separately. After the model is trained, we can apply classification by novelty detection for each class in dataset. 
%
%
Our framework is distilling randomized prior functions for data. Due to our prior functions are linear, in couple with bootstrap methods it provides a Bayes posterior~\cite{DBLP:conf/nips/OsbandAC18}. 
We tested our architectures on tasks with different amounts of data on MNIST and Omniglot datasets.

The remaining paper will be structured in the following way.
In Section 2, we describe distillation learning techniques and their application.
In Sections 3--5, we propose our method for linear network training and gradually describe the components of our architecture. In Section 3, we describe the \emph{teacher network}, in Section 4 we describe the student network, in Section 5 we describe how to train them together.
In Section 6, we present a description and results of empirical evaluations.
In Section 7, we analyze the model performance and provide an interpretation of the results.
Section 8 concludes the paper and outlines further research.






\section{Distillation}
Knowledge distillation (KD) is a method of transferring ``knowledge'' from one machine learning model called \textit{teacher} to another one called \textit{student}. 
The idea behind KD is that a \textit{teacher} network is a high-capacity model with desired high performance and a \textit{student} network is a more lightweight model \cite{DBLP:conf/nips/BaC14,DBLP:journals/corr/RusuCGDKPMKH15,DBLP:conf/iclr/UrbanGKAWMPRC17}. 
A \textit{student} cannot match the \textit{teacher}, but the distillation process brings the \textit{student} closer to the predictive power of the \textit{teacher}. 
Distillation idea was brought to the neural network community by Hinton et. al.~\cite{DBLP:journals/corr/HintonVD15}.  

In distillation learning, knowledge is transferred by training a \textit{student} model using a soft target distribution for comparison with the output layer. 
This distribution is produced by a cumbersome model with a high temperature in its output softmax
\begin{equation}
    S_{i}=\frac{\exp \left(z_{i} / T\right)}{\sum_{j} \exp \left(z_{j} / T\right)}
\end{equation}
with $z_{i}$ are logits and $T$ is temperature. 
Another scenario of knowledge distillation training is transferring knowledge from an ensemble of highly regularized models to a smaller model~\cite{DBLP:journals/corr/HintonVD15}. 

Distillation can also be applied for adversarial permutation~\cite{DBLP:conf/sp/PapernotM0JS16}, born-again neural networks~\cite{DBLP:conf/icml/FurlanelloLTIA18} and Global Additive Explanations~\cite{DBLP:conf/nips/tan18}. Furthermore, Sau and Balasubramanian~\cite{DBLP:journals/corr/SauB16} proposed to add random perturbations into soft labels for simulating learning from multiple \textit{teachers}.

Surprisingly, the distillation method often allows smaller \textit{student} network to be trained to mimic the larger and deeper models very accurately, while the \textit{student} trained on the one-hot hard targets cannot achieve the same results. The clear reason for this awaits to be discovered.

Random Network Distillation~\cite{DBLP:journals/corr/abs-1810-12894} was used for exploration in environments with sparse rewards.
In this setting, distillation allows an agent to determine whether states were visited or not, and therefore use curiosity for the exploitation. 
Let $\mathcal{O}$ be a set of observable states. 
\textit{Predictor} is a network $\hat{f} : \mathcal{O} \rightarrow \mathbb{R}^{k}$ that is trained to predict behavior of \textit{target} $f : \mathcal{O} \rightarrow \mathbb{R}^{k}$ during interaction with the environment using MSE $\|\hat{f}(\mathrm{x} ; \theta)-f(\mathrm{x})\|^{2}$ for updating parameters $\theta_{\hat{f}}$. 
If the difference between the predictions of the random network and the predictor at some environment state $S$ is large, the agent receives a higher curiosity reward. 
This can be considered as a model of novelty detection~\cite{DBLP:journals/sigpro/PimentelCCT14}, the training process of which is performed via distillation of a random network.
An important claim is that $\hat{f}$ can simulate the behavior of $f$ if their expressive powers are identical \cite{DBLP:journals/corr/abs-1810-12894}. We conducted this property and downed the expressive of networks to the linear.

\section{Class-Dependent Teacher Distillation}

Consider a classification problem with object set $X = \mathbb{R}^{d}$ and label set $Y=\{1,\ldots, C\}.$ 
We are given a labelled dataset $D = \left\{x^{i}, y^{i}\right\}_{i=1}^{n},$ where $x^{i} \in X$ and $y^{i} \in Y$.
Assume that the classification is performed by distance-based learning with some classifier $A : \mathbb{R}^{k} \to Y$ that works with object representation is some $k$-dimensional space. 
Consider also some \textit{target} function $Q_{\phi} : X \rightarrow \mathbb{R}^{k},$ which maps objects to this vector space, which is can be represented with a LNN or just a matrix.


Our idea is to create linear \textit{predictor} $P_{\theta_{c}}: X \rightarrow \mathbb{R}^{k}$ for each class $c$ that would simulate behavior of the \textit{target} function on this class.
Each \textit{predictor} is trained to map objects $x^{i}_{c}$ of class $c$ into representation $Q_\phi(x^{i}_{c})$.

Due to functions $\left\{P_{\theta_{c}}\right\}$ and $Q_{\phi}$ are linear, we can denote them in a matrix form. 
For example, $Q_{\phi}$ is can be considered as a matrix multiplication $W_{l} W_{l-1} \cdots W_{1}$. 
During training, labels $ y^{i} \in\{0,C\}$ are used to activate one of the \textit{predictors} $\left\{P_{\theta_{c}}\right\}_{c=1}^{C}$. The process of training uses MSE:
\begin{equation}
    \mathcal{L}\left(P_{\theta_{c}}\right)=\frac{1}{N}\sum_{i=1}^{N}\| P_{\theta_{c}} x^{i}_{c} - W_{l} W_{l-1} \cdots W_{1} x^{i}_{c}\|^{2}
\end{equation}

At the model evaluation step, we make prediction using the distance between $P_{\theta_{c}}(x^{i}_{c})$ and $Q_\phi(x^{i}_{c})$ for each class $c$: resulting label is chosen as $\arg\min_{c}\| P_{\theta_{c}}(x) - Q_\phi(x)\|^{2}.$

It is important to note, that despite each of $\left\{P_{\theta_{c}}\right\}$ is linear, the composition of their results cannot be expressed with a linear function. 
Nevertheless, on each step of learning process, both \textit{teacher} and \textit{student} are linear.

We replaced the classification problem with the problem of approximating a linear function with linear functions associated with classes and call this method \textbf{One to Many Distillation}~(O2MD).

\subsection{Bayesian Interpretation}

Following the analogies in RND, our distillation framework can be presented as randomized prior functions for data $D= \left\{x^{i}, y^{i}\right\}_{i=1}^{n}$. Osband et al.showed~\cite{DBLP:conf/nips/OsbandAC18} that bootstrap approaches~\cite{DBLP:journals/siamrev/Sivaganesan94a} and randomized prior functions provide a Bayes posterior in the linear case and allow for provides much cheap computing in comparison with Exact Bayes. 

In this setting, we investigate a distribution over functions $G_{\theta} = P_{\theta} + Q_{\phi},$ where parameters $\theta$ are specified by minimizing the expected prediction error with regularization $\mathcal{R}(\theta)$~\cite{DBLP:conf/nips/OsbandAC18}. In the formalism we propose, we have specific distribution $G_{\theta_{C}}$ for each class $c$:
\begin{equation}
\theta_{c}=\underset{\theta_{c}}{\arg \min } \mathbb{E}_{\left(x^{i}, y^{i}\right) \sim D}\left\|(P_{\theta_{c}}+Q_{\phi})\left(x^{i}_{c}\right)-y^{i}\right\|^{2}+\mathcal{R}(\theta_{c}).
\end{equation}
Parameters $\phi$ are drawn from prior $q(\phi)$ over the parameters of mapping $Q_{\phi}.$ After updating on the evidence we can extract from the posterior. 
In our case, if we set $y^{i}$ equal to $0$ for every class, according to RND each distillation error is a quantification of uncertainty in predicting the constant zero function  
\begin{equation}
\theta_{c}=\arg \min _{\theta_{c}} \mathbb{E}_{\left(x_{i}, y_{i}\right) \sim D}\left\|P_{\theta_{c}}\left(x^{i}_{c}\right)+Q_{\phi}\left(x^{i}_{c}\right)\right\|^{2}.
\end{equation}

By default, we interpret our model as an unbiased ensemble with shared parameters, but in practice we can actually consider the model as ensemble of \textit{target-predictor} networks for each class. 
In this settings, predictions and target in each ensemble are taken as the sum of \textit{target} and \textit{predictor} functions. 

During bootstrapping with zero target, ensemble without priors has almost zero predictive uncertainty as $x$ becomes large and negative~\cite{DBLP:conf/nips/OsbandBPR16}, which leads to to arbitrarily poor decisions~\cite{DBLP:journals/corr/OsbandR15}.


\begin{figure*}[h!]\centering
\vskip 0.2in
\begin{center}
\centerline{\includegraphics[width=13.0cm]{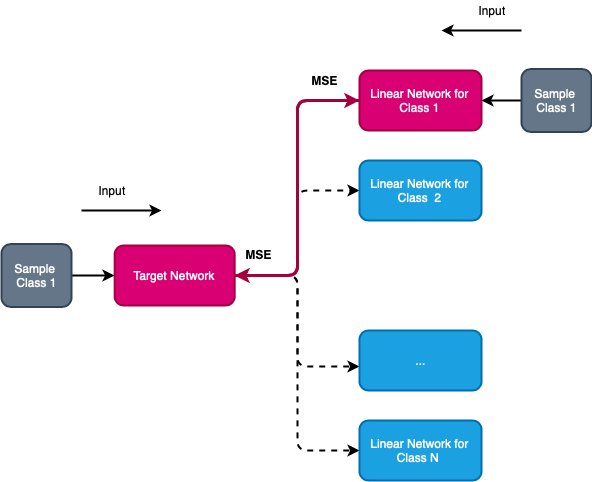}} 
\caption{Bidirectional Distillation Architecture. By red denoted: Activated predictors, which are updated using MSE between it and Target outputs. The target network is connected by dashed lines with predictors, which will be activated for training, if the appropriate class is specified at the input.}
\label{fig:1}
\end{center}
\vskip -0.2in
\end{figure*}

\section{Class-Dependent Students Distillation}

In the previous Section, we have presented approach for learning linear \textit{predictors} with linear \textit{target}. 
In this Section, we present two approaches for selecting this target. 

First, we will formulate desirable properties of \textit{target}.
Due to the \textit{predictors} simulate \textit{target} on corresponding classes, when we compare outputs, we can more clearly distinguish one of the \textit{predictor} if dissimilarity with other classes is much larger. 
For example, if the \textit{target} output for class $1$ is very different from the output for other classes, the trained \textit{predictor} $P_{\theta_{1}}$ for this class will be closer to the target than other ones. 
It may be possible if the \textit{target} outputs for each class are far from each other. 

One of the way to choose $Q_{\phi}$ that maps classes in distinct regions is to train it directly to do this. It can be done, for example, by updating our \textit{target} $\phi$ parameters using distillation from some \textit{teacher}. 

In our case, we noticed an interesting feature of distillation: \textit{teacher} does not have to be able to generalize data. 
It would be sufficient if \textit{teacher} function just maps the train data classes in different regions of its output space. 
In contrast to the standard paradigm of distillation, our \textit{teacher} is a set of random initialized linear functions $\left\{P_{\zeta_{c}}\right\}_{c=1}^{C}$ for each class $c$ that map data in different spaces by default. 
At step $i$ we choose a linear neural network from $\left\{P_{\zeta_{c}}\right\}_{c=1}^{C}$ corresponding to label $y^{i}$ and use its transformation as a target for our $Q_{\phi}$. Learning/optimization is performed by minimizing loss
\begin{equation}
    \mathcal{L}\left(Q_{\phi}\right)=\frac{1}{N}\sum_{i=1}^{N}\| W_{l} W_{l-1} \cdots W_{1} x^{i}_{c}- P_{\zeta_{c}} x^{i}_{c}\|^{2}
\end{equation}

We call this method \textbf{Many to One Random Distillation}~(M2ORD). The prediction accuracy of the model trained in this way is sub-optimal. It seems to be difficult to learn data distribution from the \textit{teachers}' outputs because the transformation the \textit{teachers} apply to the data is not linear. 

\subsection{Orthonormal initialization} 
The idea of orthogonalization is to make $\left\{P_{\zeta_{c}}\right\}_{c=1}^{C}$ functions orthogonal to each other to maintain distinct mapping for each class. We use the Gram–Schmidt process~\cite{DBLP:journals/computing/Hoffmann89} to make \textit{teacher} functions orthogonal. This method is used in linear algebra for orthonormalizing a set of vectors. We create $\left\{P_{\zeta_{c}}\right\}_{c=1}^{C}$ for each class with size of $(n,d)$ and join them in single matrix $A$ size of $(C \cdot n, d)$, where $d > C\cdot n$.

We use singular vector decomposition (SVD)~\cite{DBLP:journals/cacm/BusingerG69} to obtain an orthogonal matrix. SVD of a matrix $A$ is the factorization of $A$ into the product of three matrices: $A=U D V^{T},$ where $U$ is orthogonal. After obtaining results of SVD, we take $U$ and split it into $C$ matrices $\left\{P_{\zeta_{c}}\right\}_{c=1}^{C}$. Each such matrix $P_{\zeta_{c}}$ consists of strings in $U$ corresponding to the objects of class $c$. As a result, all rows of all \textit{teacher} matrices are linearly independent and project points into subspaces that are orthogonal: 
\begin{equation*}
    \left\|\mathrm{P_{\zeta_{1}}}\right\|=\left\|\mathrm{{P_{\zeta_{1}}}}\right\|=\dots=\left\|\mathrm{{P_{\zeta_{c}}}}\right\|=1
\end{equation*}
\begin{equation*}
    {P_{\zeta_{1}}} \perp {P_{\zeta_{1}}} \perp \dots \perp {P_{\zeta_{c}}}
\end{equation*}

\section{Bidirectional Distillation}

In the previous two Section, we presented two ideas, namely, O2MD for learning linear \text{predictor} functions using distillation and M2ORD for learning linear \text{target} function using distillation. In this Section, we combine these two ideas in a method we call \textbf{Bidirectional Distillation}. 

After initialization, our predictors $\left\{P_{\theta_{c}}\right\}_{c=1}^{C}$ are equal to $\left\{P_{\zeta_{c}}\right\}_{c=1}^{C}$, so at the first step we pretrain our function $Q_{\phi}$ by $\left\{P_{\theta_{c}}\right\}_{c=1}^{C}$ in M2ORD and then distill this knowledge back into $\left\{P_{\theta_{c}}\right\}_{c=1}^{C}$ by O2MD training, 

The learning procedure is presented in Figure 1. During Bidirectional Distillation, we alternate O2MD and M2ORD in different proportions, we consistently train them a certain number of iterations, allowing to be updated several times in each epoch.

\section{Experiments and Results}



In this Section, we describe the results of experiments, comparing One-To-Many and Bidirectional variants of a LDL model against a deep fully connected neural network, logistic regression and na\"ive linear model. In na\"ive settings, our \textit{predictors} are trained without \textit{target}. Each \textit{predictor} is trained to produce output representation as close as possible to the original input. At the prediction stage, we only measure the distance between each \textit{predictor} output and sample $x_{i}$. 

All of our experiments are formulated within few-shot problem framework. Each model was provided with a set of $m$ labelled samples from $C$ classes. In the few-shot learning terminology, $C$ is traditionally called a \emph{way} and $m$ is called a \emph{shot}. The task is $C$-class classification. Unlike traditional few-shot learning models, our approach does not imply knowledge transfer between episodes, being thus more similar to small-sample learning~\cite{DBLP:journals/corr/abs-1808-04572}. To avoid bias between small number of samples and reported results, each model was trained for 100 independent trials and average accuracy was reported.
We ran experiments on two datasets with images of low resolutions, namely MNIST and Omniglot. 
Our experiments were conducted on a machine with two separate NVIDIA GeForce GTX 1080 Ti, 4 core Intel i5-6600K processors and 47.1 GB RAM memory. We used PyTorch 1.1 with CUDA 10 support for implementation of all models presented in the study. All datasets were uploaded into memory preliminary for faster computation.



\subsection{MNIST}

For experiments on the MNIST dataset, a small number of samples from each of the digits were given to the model, thus the way was always set to 10. After a model was trained on the given samples, we calculated accuracy on the whole official test part of the MNIST dataset for the easier comparison with well-known approaches. Models were trained on the shots of sizes 1, 10, 50, 200, 300. For testing the models, we performed no data preprocessing and image augmentation. As a result, a single sample was a flattened representation of the $28\times 28$ grayscale image. We also chose learning rates between $1e-3$, $1e-4$ and $5e-5$ values. We noticed that a large numbers of epochs were not sufficient for the linear setup and chose the total number epochs to be not greater than 10 for all training shots. 

\begin{table}[h!]\centering
\caption{Results on MNIST Dataset}
\label{tab:1}
\vskip 0.15in
\begin{center}
\begin{small}
\begin{sc}
\begin{tabular}{llllll}
\toprule
Num & Naive & LR & MLP & O2MD & BD \\ 
\midrule
1                   &0.127                    & 0.316                                & \textbf{0.448}                                              & 0.426                             & 0.436        \\
10                  &0.777                    & 0.679                                & 0.749                                              & \textbf{0.801}                             & 0.800          \\
50                  &0.903                    & 0.839                                & 0.881                                              & 0.912                             & \textbf{0.917}        \\
100                 &0.898                    & 0.870                                & 0.926                                              & \textbf{0.934}                             & 0.871        \\
200                 &0.942                    & 0.892                                & 0.929                                              & \textbf{0.953}                             & \textbf{0.953}        \\ 
\bottomrule
\end{tabular}
\end{sc}
\end{small}
\end{center}
\vskip -0.1in
\end{table}

Linear distillation approaches were compared with logistic regression and fully connected network. Fully connected network was chosen between configurations of one and two hidden layers of sizes 64, 256 or 1024. The most promising results were reported for each of the shot value. Results comparing the One-To-Many and Bidirectional Distillation models for different number of shots are shown in Table~1. 
It is no surprise that the accuracy increased with the size of the training dataset, but it slowed down when the size of the dataset reached the size of about thousand samples.

An important property of our architecture is that it is learned fairly quickly on a small training epochs. The convergence of each student to the teacher is presented in Figure 3, where students are $\left\{P_{\theta_{c}}\right\}_{c=1}^{C}$, and the teacher is $Q_{\phi}$ network.

\begin{figure*}[h!]
    \centering
    \includegraphics[width=12.0cm]{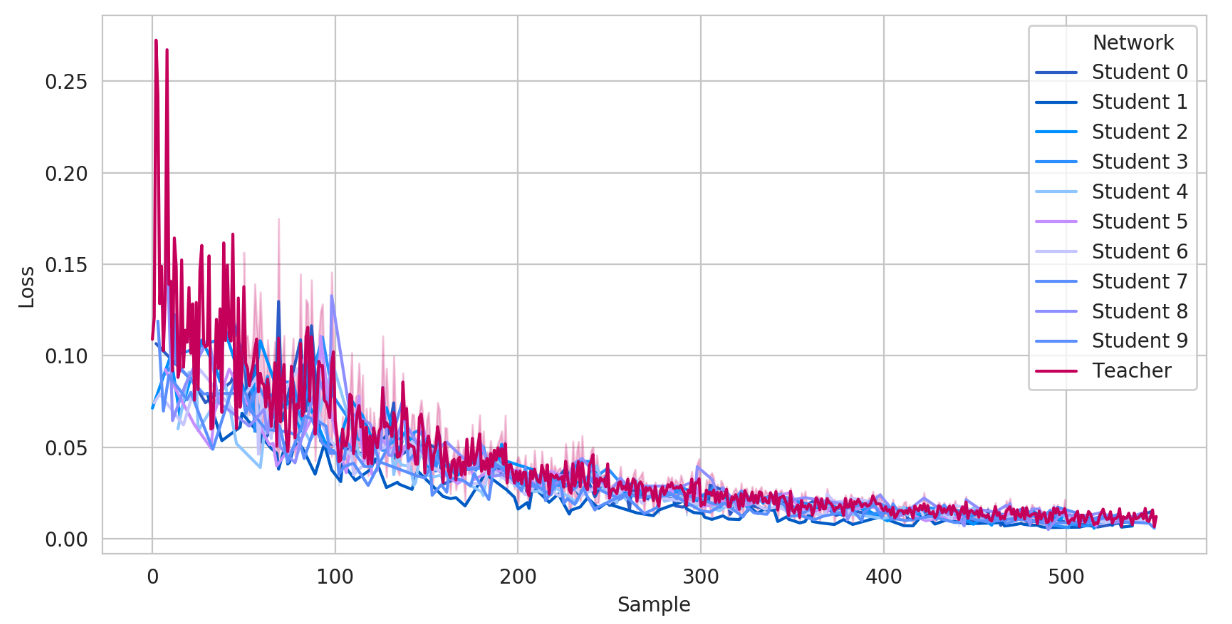}%
    \caption{Test loss curve for Bidirectional Distillation model. The model is trained for 10 epochs on MNIST dataset with 5 samples per class}%
    \label{fig:example}%
\end{figure*}

\begin{figure*}[h!]
    \centering
    \includegraphics[width=13.0cm]{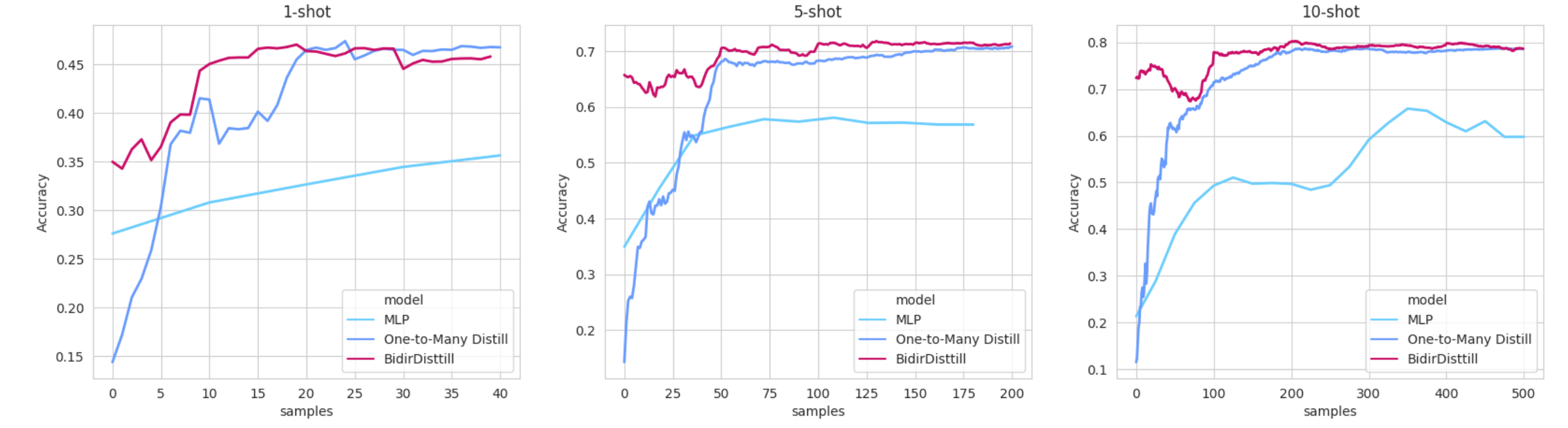}%
    \caption{Accuracy curve for the MLP, One-tp-Many and Bidirectional Distillation models. The models are trained for 10 epochs on MNIST dataset with 1, 5 and 10 samples per class}%
    \label{fig:example}%
\end{figure*}

Training a Bidirectional Distillation model and a O2MD version has advantages over the classic deep fully connected network on a small amount of data. Bidirectional training allows much faster converging to an almost best capabilities of the model after the first epochs, because the \textit{target} pre-trained on \textit{predictors} simplifies training. An example of the training process comparing MLP with two hidden layers and the distillation models is depicted in Figure 4. The $y$ axis shows accuracy on the full test part of the dataset measured after each epoch/sample.

\subsection{Omniglot}
The Omniglot dataset consists of 1623 characters from 50 different alphabets, where each of the characters was drawn by 20 different people. Following the authors of Matching Networks~\cite{DBLP:conf/nips/VinyalsBLKW16}, we augmented existing classes with rotations by multiples of 90 degrees and used 1200 characters from training and the remaining character classes for evaluation. We resized images to the size of $28 \times 28$ pixels and obtained the very same model settings as for the previous dataset.
\begin{table}[h!]\centering
\caption{Results on Omniglot Dataset}
\label{tab:2}
\vskip 0.15in
\begin{center}
\begin{small}
\begin{sc}
\begin{tabular}{llllll}
\toprule
way                 & shot & \multicolumn{2}{l}{O2MD}              & \multicolumn{2}{l}{BidirDistill}              \\ 
                    &      & \multicolumn{2}{l}{Output Size} & \multicolumn{2}{l}{Output Size} \\ 
\midrule                   
                    &      & 784                & 2000             & 784                & 2000             \\ \hline
3                   & 1    & 0.563              & 0.593            & 0.566              & 0.726            \\ 
                    & 3    & 0.683              & 0.720            & 0.726              & 0.686            \\ 
                    & 5    & 0.800              & 0.803            & 0.773              & 0.760            \\ 
                    & 10   & 0.797              & 0.830            & 0.780              & \textbf{0.880}            \\ \hline
5                   & 1    & 0.428              & 0.454            & 0.372              & 0.420            \\ 
                    & 3    & 0.614              & 0.626            & 0.560              & 0.636            \\ 
                    & 5    & 0.666              & 0.674            & 0.632              & 0.672            \\ 
                    & 10   &\textbf{0.806}              & 0.778            & 0.684              & 0.736            \\ \hline
10                  & 1    & 0.305              & 0.342            & 0.301              & 0.343            \\ 
                    & 3    & 0.480              & 0.536            & 0.457              & 0.491            \\ 
                    & 5    & 0.572              & 0.585            & 0.526              & 0.605            \\ 
                    & 10   & 0.685              & 0.689            & 0.663              & \textbf{0.697}            \\ 
\bottomrule
\end{tabular}
\end{sc}
\end{small}
\end{center}
\vskip -0.1in
\end{table}

We experimented with different learning rates ($1e-3$, $1e-4$, $1e-5$) and optimizer types (sgd, adam, adadelta) and reported results from most promising configurations. We tested our models on the way $N$ of sizes 3, 5 and 10 with shots $m$ of size 1, 3, 5, and 10. After training on the given $N\cdot m$ samples, we measured the accuracy on the $N$ unseen samples from these $N$ classes (one sample per class). For the One-To-Many and Bidirectional setups, we provided results for 784 and 2000 networks' output dimensions. To provide unbiased results, we averaged results over 100 independent runs. Results comparing studied models for different number of shots are shown in Table 2. 

Our model showed acceptable results mainly on a small number of ways. It was sensitive to hyperparameter settings, but nevertheless, was able to learn classification with proper configuration.


\section{Results Interpretation}
Lipton notes~\cite{Lipton} provide a definition of interpretability into two categories. The first relates to \textit{transparency}, i.e., ``How does the model work?'' and \textit{post-hoc} explanations, i.e., ``What else can the model tell me?'' \textit{Transparency} connotes some sense of understanding the mechanism by which the model works. The \textit{post-hoc} interpretation confer useful information for practitioners and end-users of machine learning model.
 
Our model meets these two requirements. In special cases, diagnostics of model predictions allows to fully understand model behavior, and propose hypotheses on how to improve the model performance. To demonstrate the interpretability of our model, we visualize the regions on which model is concentrated at different classes on MNIST dataset using LIME~\cite{Lime} technique that is a model interpretation technique based on attempting to understand a model by perturbing the input of data samples and observing how its predictions change. This test answering the questions ``Why the model made this prediction?'' and ``Which variables influenced the prediction?'' These relations are clearly shown in our model.

Complex models such as ensemble methods or deep networks are not easy to understand. To interpret it, we must use a simpler model, which we set as any interpretable approximation of the original model. Dense embeddings are not interpretable, and applying LIME probably will not improve interpretability. In our case, the best explanation of the model is the model itself; it perfectly represents itself and is easy to understand, see Figure 4.
\begin{figure}[h]
    \centering
    \includegraphics[width=13.0cm]{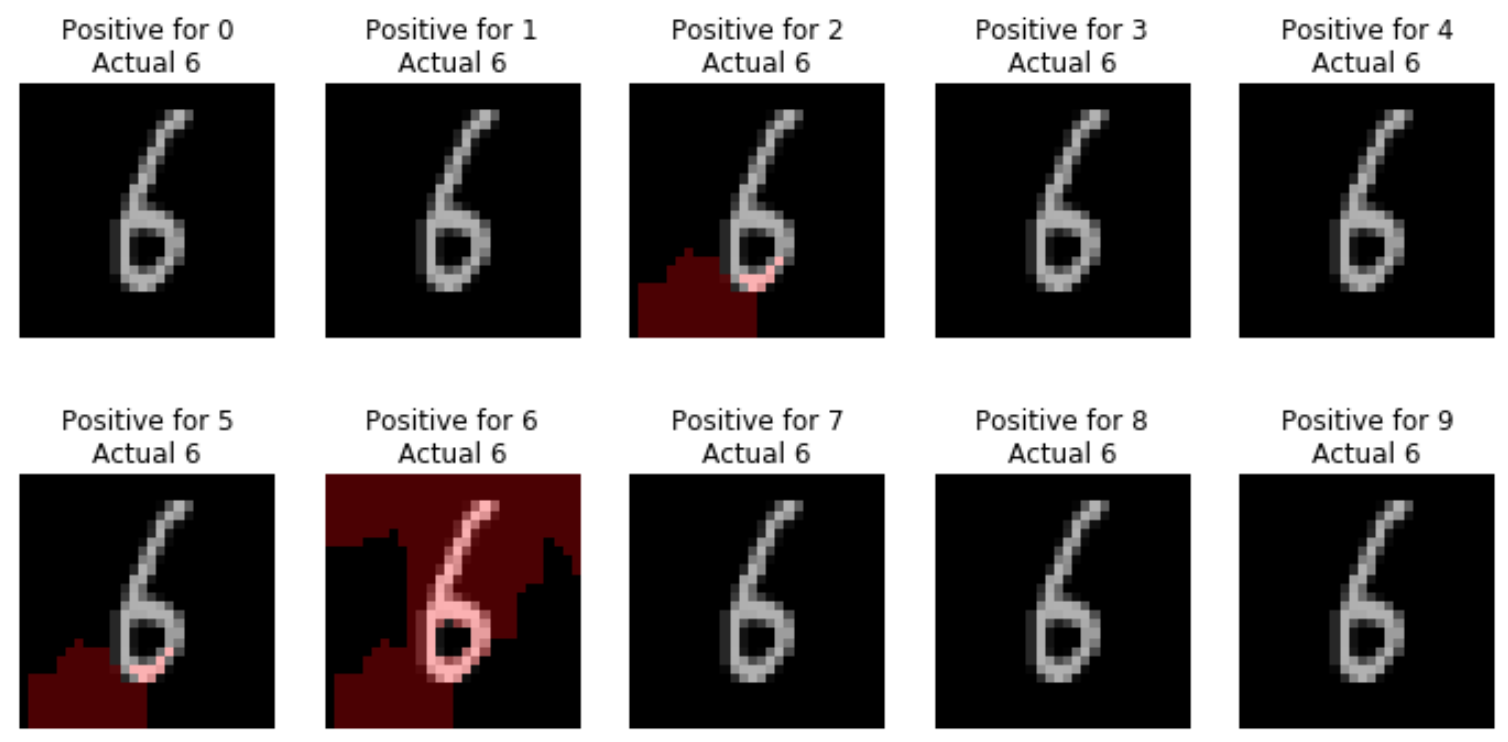}%
    \caption{Model correct predictions interpretation on MNIST dataset}%
    \label{fig:example}%
\end{figure}

LDL outperforms other methods on small datasets since it is not that important how overfitted a single predictor is if other predictors are overfitted too. At the evaluation stage, we compare each predictor with others. After a few training samples, each predictor starts to recognize one class better than others, and it is enough to start classifier. The model performance is strongly correlated with the power of each predictor, if predictor's power is imbalanced, the model will more often choose the prediction of a stronger predictor. To demonstrate this, we visualized model incorrect predictions in Figure 5.

\begin{figure}[h]
    \centering
    \includegraphics[width=13.0cm]{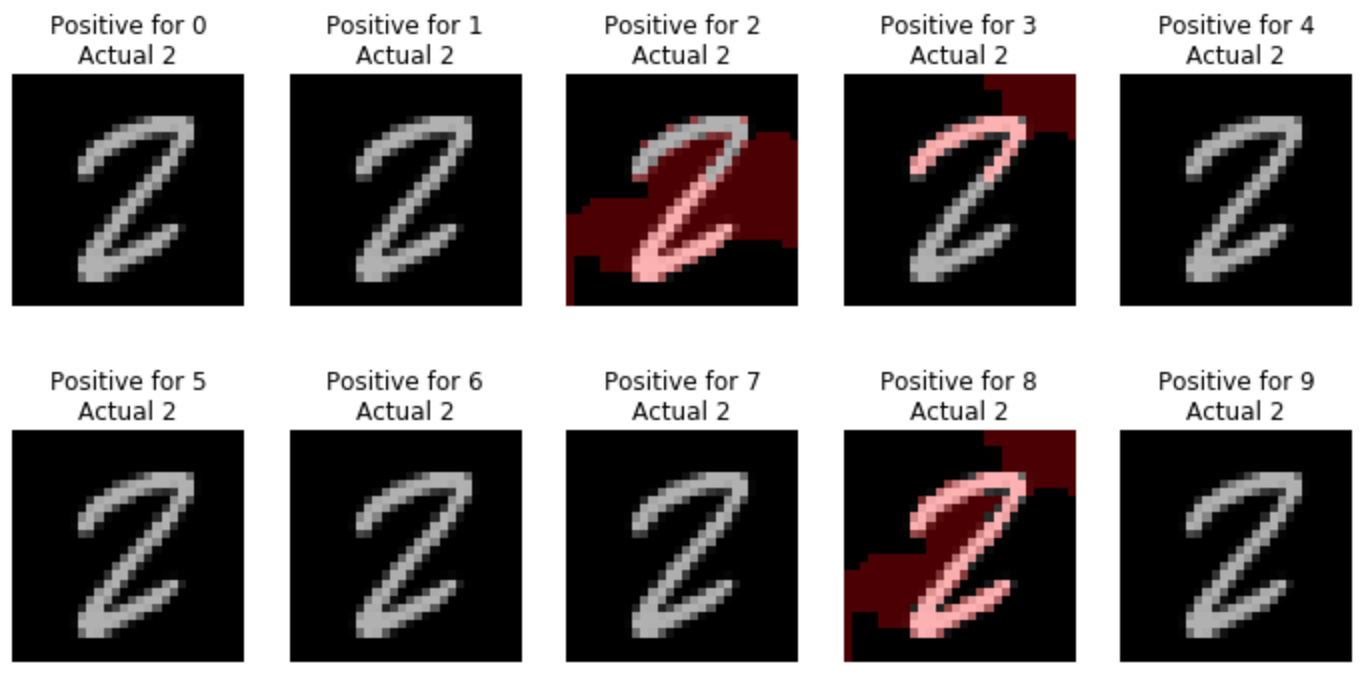}%
    \caption{Model wrong predictions interpretation on MNIST dataset}%
    \label{fig:example}%
\end{figure}

As it seems, the model made an incorrect prediction, because of the imbalance between predictor for class '2'and '8'. Simply to avoid this problem, we trained predictor for class '2' in MNIST on one more epoch than all other predictors. As a result, after retaining our model made the correct prediction on this sample from the test dataset, see Figure 6.

\begin{figure}[h]
    \centering
    \includegraphics[width=13.0cm]{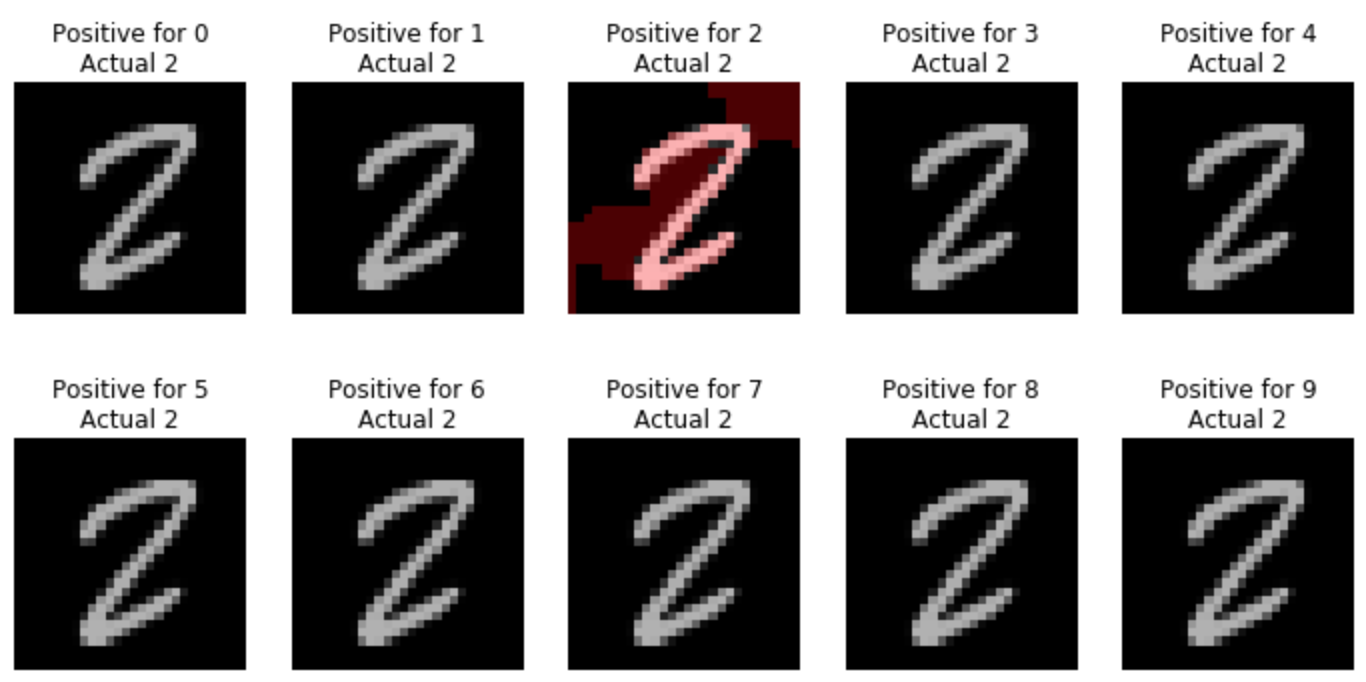}%
    \caption{Model predictions interpretation on previously wrong sample on MNIST dataset}%
    \label{fig:example}%
\end{figure}

\section{Conclusion and Future Work}

In this paper, we present an architecture based on several methods of random function distillation using linear neural networks. The first method is the training of a single network to simulate the behavior of another on a particular class. The second method is the training of a single network to predict the behavior of many other networks for each class in dataset. Our architecture can be considered through the Bayesian lens too. The motivation for our work was to create an architecture, which consists of linear functions capable of classifying on a small dataset. We tested our model on several datasets and showed results comparable to the results of nonlinear models on small amounts of data.

For the Omniglot dataset, we tested our architecture in the few-shot learning paradigm. Our model does not have the key concept of the paradigm to preserve knowledge in the learning process between sub-datasets (episodes). This is an important issue for future research, so one direction of the further studies can be usage of our method in the few-shot tasks on a higher level. There is abundant room for further progress in using distillation as a method of learning and explore open questions in this area. 

\section{Acknowledgments}
We would like to thank Artem Zholus for productive discussions.

\end{document}